\documentclass[letterpaper, 10 pt, conference]{ieeeconf}  

\IEEEoverridecommandlockouts                              

\title{\LARGE \bf
OG-Gaussian: Occupancy Based Street Gaussians for Autonomous Driving
}

\author{Yedong Shen$^{1}$, Xinran Zhang$^{1}$, Yifan Duan$^{1}$, Shiqi Zhang$^{2}$, Heng Li$^{1}$, Yilong Wu$^{1}$\\ Jianmin Ji$^{1}$ and Yanyong Zhang$^{2}$*~\IEEEmembership{Fellow,~IEEE}
\thanks{*The corresponding author.}
\thanks{$^{1}$~School of Computer Science and Technology, University of Science and Technology of China, Hefei 230026, China {\tt\small \{sydong2002, zxrr, dyf0202, li\_heng, elonwu\}@mail.ustc.edu.cn, jianmin@ustc.edu.cn}}
\thanks{$^{2}$~School of Artificial  Intelligence and Data Science, University of Science and Technology of China, Hefei 230026, China {\tt\small zhangshiqi\_1127@mail.ustc.edu.cn, yanyongz@ustc.edu.cn} }
}

\usepackage{amsmath}
\usepackage{amssymb}
\usepackage[utf8]{inputenc}

\usepackage{booktabs}  
\usepackage{graphicx}
\usepackage{multirow}
\usepackage{xcolor}
\makeatletter
\let\NAT@parse\undefined
\makeatother
\usepackage[colorlinks,
            urlcolor=blue,
            linkcolor=blue,       
            anchorcolor=blue,  
            citecolor=green        
            ]{hyperref}
\begin{document}

\maketitle
\thispagestyle{empty}
\pagestyle{empty}

\begin{abstract}
Accurate and realistic 3D scene reconstruction enables the lifelike creation of autonomous driving simulation environments. With advancements in 3D Gaussian Splatting (3DGS), previous studies have applied it to reconstruct complex dynamic driving scenes. These methods typically require expensive LiDAR sensors and pre-annotated datasets of dynamic objects. To address these challenges, we propose OG-Gaussian, a novel approach that replaces LiDAR point clouds with Occupancy Grids (OGs) generated from surround-view camera images using Occupancy Prediction Network (ONet). Our method leverages the semantic information in OGs to separate dynamic vehicles from static street background, converting these grids into two distinct sets of initial point clouds for reconstructing both static and dynamic objects. Additionally, we estimate the trajectories and poses of dynamic objects through a learning-based approach, eliminating the need for complex manual annotations. Experiments on Waymo Open dataset demonstrate that OG-Gaussian is on par with the current state-of-the-art in terms of reconstruction quality and rendering speed, achieving an average PSNR of 35.13 and a rendering speed of 143 FPS, while significantly reducing computational costs and economic overhead.
\end{abstract}

\section{INTRODUCTION}
Reconstructing realistic, geometrically accurate 3D scenes has long been a key objective in computer vision. Thanks to advancements in technologies like Neural Radiance Fields (NeRF)~\cite{mildenhall2021NeRF} and 3D Gaussian Splatting (3DGS)~\cite{kerbl20233d}, generating high-precision 3D models is now more attainable. These technologies greatly enhance the realism of virtual environments and have significant applications in various fields, such as medical imaging~\cite{wang2021birds}, surgical navigation~\cite{liu2024endogaussian} and virtual reality~\cite{jiang2024vr, abdal2024gaussian}. In autonomous driving, the reconstruction technologies can provide precise 3D models of the surroundings, including street, building, and even dynamic object. This capability improves the autonomous system's navigation ability and enables simulation of extreme scenarios, expanding the boundaries of reality while digitizing it. 



\begin{figure}[h]
    \centering
    \includegraphics[width=\columnwidth]{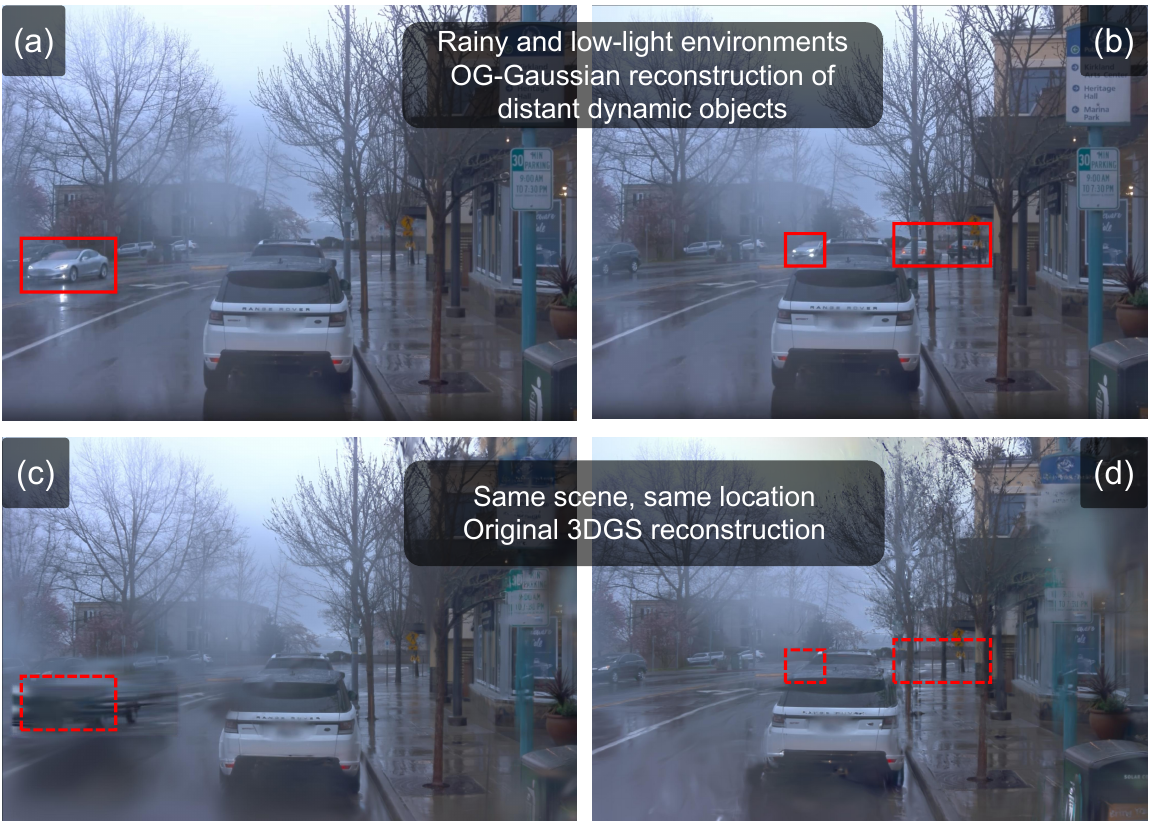} 
    \caption{\textbf{OG-Gaussian reconstruction example. }(a) and (b) visualize the performance of our method in reconstructing distant dynamic objects and the scene under rainy and low-light conditions. (c) and (d) present the reconstruction results of original 3DGS from the same viewpoint. Red bounding boxes indicate the location of dynamic objects in the ground truth.}
    \label{fig:intro} 
    \vspace{-20pt}
\end{figure}


To achieve the high-precision reconstruction of autonomous driving scenes, NeRF is used as the foundational technology, representing the scene as a continuous 3D volume through neural network~\cite{xie2023s, wu2023mars, yang2023emernerf}. While this method generates high-quality outdoor scenes, it comes with the cost of requiring extensive training resources and slower rendering speed. With the emergence of 3DGS, the low-cost, fast-rendering 3D scene reconstruction method quickly captured widespread attention. The native 3DGS is not well suited for handling large outdoor scene with dynamic objects. To adapt this technology for autonomous driving scene reconstruction, existing works of 3DGS put their attention on integrating LiDAR-generated point clouds and using annotated 3D bounding boxes to reconstruct street scenes with dynamic objects~\cite{luiten2023dynamic, zhou2024drivinggaussian, yan2024street}. They have successfully separated dynamic objects from static background, achieving promising reconstruction results under low training cost\cite{liu2025dvlo}. 




However, these techniques need expensive LiDAR to generate point clouds and datasets with pre-labeled dynamic vehicle boundaries and trajectories. To mitigate this constraint, we bring Occupancy Prediction Network(ONet)~\cite{mescheder2019occupancy}, a new approach in autonomous driving perception into the field of 3D scene reconstruction. Since ONet models the real world as voxel grids with semantic information directly, we can eliminate the need for costly LiDAR and resolve the problem of bounding boxes failing to capture unannotated objects\cite{liu2024difflow3d}. 

Based on the considerations, we propose OG-Gaussian, a new solution for autonomous driving scene reconstruction. Our method starts by capturing surround-view images using cameras mounted on the vehicle. Then, we use a Occupancy Prediction Network (ONet) to obtain Occupancy Grid (OG) information about the surrounding world. By leveraging the semantic information of the Occupancy Grid, we separate the original scene into street scene and dynamic vehicles. 

Afterward, we convert the Occupancy Grids of background street into point clouds and transform dynamic vehicle grids into an initial set of point clouds via 2D image projection. Instead of relying on expensive LiDAR point clouds as initial point clouds (prior), the point clouds obtained from Occupancy Grids can serve as low-cost alternative prior in our approach. These point clouds will be converted into optimizable sets of Gaussian ellipsoids. To track dynamic vehicles, we define their initial points' positions and rotation matrices as learnable parameters. This allows us to optimize the vehicles' poses and trajectories, which describe the operation of dynamic vehicles in the real world. In this way, our method does not require pre-labeled trajectories or bounding boxes of dynamic objects. The final Optimized Gaussian ellipsoids will be projected into 2D space to render the reconstructed autonomous driving scenes in the real world.

We conduct experiments on several representative scenes in Waymo Open dataset~\cite{sun_2020_cvpr}, which includes a wide variety of driving scenes collected from diverse urban and suburban areas. Experiments demonstrate that our method is comparable to the current state-of-the-art methods using LiDAR in reconstruction quality and rendering speed. Additionally, we carry out ablation study to confirm the effectiveness of using processed Occupancy Grids as prior in reconstructing autonomous driving scenes. We provide fast, low-cost 3D scene reconstruction method for downstream tasks.


The main contributions of this paper are as follows:
\begin{itemize}

\item We introduce OG-Gaussian, incorporating Occupancy Grid into autonomous driving scenes reconstruction. This approach eliminates the need for expensive LiDAR to generate initial point clouds, requiring only image input and significantly reducing the cost of 3D scene reconstruction.
\item We leverage the semantic properties of Occupancy Grids to separate dynamic vehicles from static background and estimate their poses, removing the need for manual labeling of dynamic objects.
\item Through extensive experiments, our method achieves an average PSNR of 35.13 and 143 FPS, without relying on LiDAR or any annotations. It's comparable to state-of-the-art methods in terms of reconstruction quality and rendering speed. 

\end{itemize}

\section{RELATED WORKS}

\noindent\textbf{Occupancy Network and applications. }Obtaining accurate semantic 3D Occupancy Grids is crucial for downstream tasks, making research on the Occupancy Prediction Network highly valuable\cite{duan2024cellmap}. MonoScene~\cite{cao2022monoscene} completes dense 3D grids from a single RGB image, it ensures spatial consistency via 2D-3D projection and 3D context prior. By extending BEV representation to 3D, TPVFormer~\cite{huang2023tri} proposes a new representation named TPV and combines it with a transformer to predict 3D semantic Occupancy Grids. SurroundOcc~\cite{wei2023surroundocc} utilizes an innovative method to generate dense occupancy labels and employed a 2D-3D attention mechanism to predict Occupancy Grids from multi-camera images. To facilitate future research, Occ3D~\cite{tian2024occ3d} introduces two benchmark datasets (Occ3D-Waymo and Occ3D-nuScenes) and a new model (CTF-Occ network). As the accuracy of occupancy prediction networks improves, many related applications are also emerging. OccWorld~\cite{zheng2023occworld} introduces a world model framework based on 3D occupancy space. It demonstrates an effective capability in modeling scene evolution on the nuScenes benchmark. OCC-VO~\cite{li2024occ} transforms 2D camera images into 3D semantic occupancy to address the depth information challenge in Visual Odometry(VO). 

\noindent\textbf{3D scenes reconstruction for Autonomous Driving. } 
Simulating real-world street scene is essential for developing and testing autonomous driving systems. A prime example is CARLA~\cite{dosovitskiy17}, a well-known open-source driving simulator that's widely used for creating complex 3D environments. But the scenes created by these simulators often lack the realism needed to fully immerse users in real-world environments. 
To adapt NeRF for unbounded and dynamic field like autonomous driving, ~\cite{martin2021NeRF, turki2022mega} improves NeRF to model multi-scale urban scenarios. ~\cite{xu2023grid} combines compact multi-resolution ground features with NeRF to achieve high-fidelity rendering. Research in ~\cite{ost2021neural, song2022towards} begin to explore handling dynamic scenes with multiple objects, leading to Suds~\cite{turki2023suds} processing scenes into static backgrounds and dynamic objects. While S-NeRF~\cite{xie2023s} introduces LiDAR as a form of supervision, MARS and EmerNeRF~\cite{wu2023mars, yang2023emernerf} further optimize NeRF for outdoor scene reconstruction to enhance its performance. However, NeRF is limited by its high training costs and slow rendering speed, so attention of industry is gradually shifting toward 3DGS due to its lower training costs and faster rendering speed~\cite{kerbl20233d}. \cite{yang2024deformable} uses a transformer to model Gaussian motion from monocular images. ~\cite{luiten2023dynamic} parametrizes the entire scene using a set of variable dynamic Gaussians. DrivingGaussian~\cite{zhou2024drivinggaussian} firstly introduces LiDAR point clouds as a prior and incrementally reconstructs the entire scene. Street Gaussians~\cite{yan2024street} reconstructs dynamic scenes with separate rendering and LiDAR. S3 Gaussian~\cite{huang2024textit} improves this using self-supervised vehicle bounding boxes. 

\begin{figure*}[h]
\centering
\includegraphics[width=0.95\textwidth]{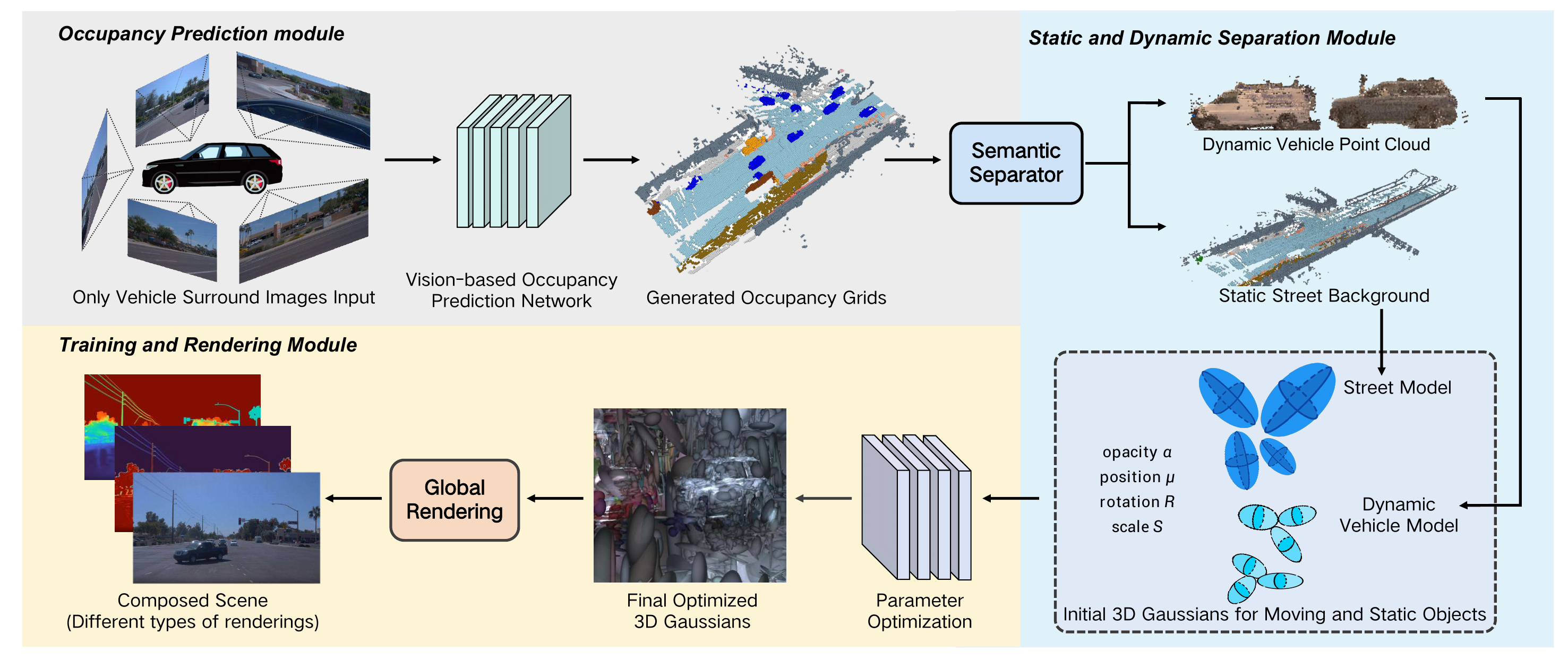}
\vspace{-3pt}
\caption{\textbf{Overview of OG-Gaussian. }OG-Gaussian utilizes a trained 3D Occupancy Prediction network to obtain Occupancy Grid data for the scene. It separates static and dynamic objects into different initial point cloud models using semantic information. After the separated reconstruction, we globally render both static and dynamic objects, producing 3D scenes, depth maps and so on.}
\label{fig:overview}
\vspace{-14pt}
\end{figure*}

\section{PRELIMINARY}

Firstly, we will introduce the classical 3DGS algorithm and 3D Occupancy Grid in the preliminary section.

\noindent\textbf{Classical 3DGS algorithm. }In the first step of reconstruction, we initialize the 3DGS point clouds by placing a Gaussian at each point, other parameters randomly initialized except the center point position. COLMAP~\cite{schoenberger2016sfm} is used to generate the Structure from Motion (SfM) point clouds as the prior. After initialization, each Gaussian ellipsoid can be represented by Eq.~\ref{1}:
\begin{equation}
    \begin{aligned}
     G(x)&=e^{-\frac{1}{2}(x)^{T} \Sigma^{-1}(x)} \\
     \Sigma&=R S S^{T} R^{T}
    \end{aligned}
    \label{1}
\end{equation}

$G(x)$ is the probability distribution function of the Gaussian, where $x$ is a three-dimensional vector representing a point in space. $\Sigma$ is the covariance matrix that describes the shape, orientation, and size of the Gaussian in 3D space. This can be written in $R S S^{T} R^{T}$, so the covariance matrix can be constructed by rotation and scaling matrix (R and S).

After representing the 3D Gaussians, we can rasterize the rendering by $ \alpha 
 $ blending\cite{wu2024mm}. From near to far, each ellipsoid projects onto the image plane, and the overlapping areas can be fused through rasterization to generate a new image. Eq.~\ref{2} is the loss function used in 3D Gaussian, where $\mathcal{L}_{1}$ is used to measure the pixels’ difference and $\mathcal{L}_{\text {D-SSIM }}$ is used to measure the structural similarity between two images.
 
 \begin{equation}
      \mathcal{L}=(1-\lambda) \mathcal{L}_{1}+\lambda \mathcal{L}_{\text {D-SSIM }}
      \label{2}
 \end{equation}

The properties of the 3D Gaussian sphere can be updated through gradient back propagation, the cloning and splitting of the Gaussian spheres can also be controlled. The gradient back propagation is divided into two branches, the upper branch updates the properties of the 3D Gaussian ellipsoid, and the lower branch realizes the cloning and splitting of the 3D Gaussian ellipsoid. 


\noindent\textbf{3D Occupancy Grid.} Occupancy Prediction Network~\cite{mescheder2019occupancy} generates 3D grids $V\in\mathbb{R}^{H\times W\times D}$, where $H, W, D$ represent the height, width, and depth of the grid. Each element $V(i, j, k)$ in the grid corresponds to the grid cell at position $(i, j, k)$. A threshold $\tau$ is applied to classify whether each grid is occupied or not: 

\begin{equation}    
  V(i, j, k)=\begin{cases}
  1&\mathrm{if~}p(i, j, k)\geq\tau\\
  0&\mathrm{if~}p(i, j, k)<\tau
  \end{cases}
  \label{3}
\end{equation}

Assigning $N$ semantic categories to the grid (such as buildings, roads, and vehicles), the state of grid can be represented by $V(i, j, k)$ and the semantic category $C(i, j, k)$. $C(i, j, k)$ can be acquired by choosing the category $c$ with the highest probability using Eq.~\ref{4}: 

\begin{equation}
    C(i, j, k)=\arg\max_c[p_1(i, j, k)\ldots p_N(i, j, k)]
    \label{4} 
\end{equation}

In this way, we can predict the occupancy state of each grid and get detailed information about its semantic category.

\section{METHOD} \label{sec:method}
Next, we will provide a detailed explanation of the OG-Gaussian method, as shown in Fig.~\ref{fig:overview}


\subsection{OG-Gaussian}
In this section we focus on the basic structure of the OG-Gaussian and introduce how we represent street scene and dynamic vehicles with two different sets of  point clouds. We will provide a detailed explanation of them below.

\noindent\textbf{Street model. }The initial point clouds of the Street model is a set of points in the world coordinate system. According to the information in the preliminary, the parameters of 3D Gaussians can be expressed as the covariance matrix $\Sigma_{s}$ and the position vector $\mu_{s} \in \mathbb{R}^3$. The matrix $\Sigma_{s}$ can be split into the rotation matrix $R_s$ and the scaling matrix $S_s$, this recovery process is:

\begin{equation}
    \Sigma_{s} = R_sS_sS_s^TR_s^T
    \label{5}
\end{equation}

\begin{figure}[h]
    \centering
    \includegraphics[width=0.85\columnwidth]{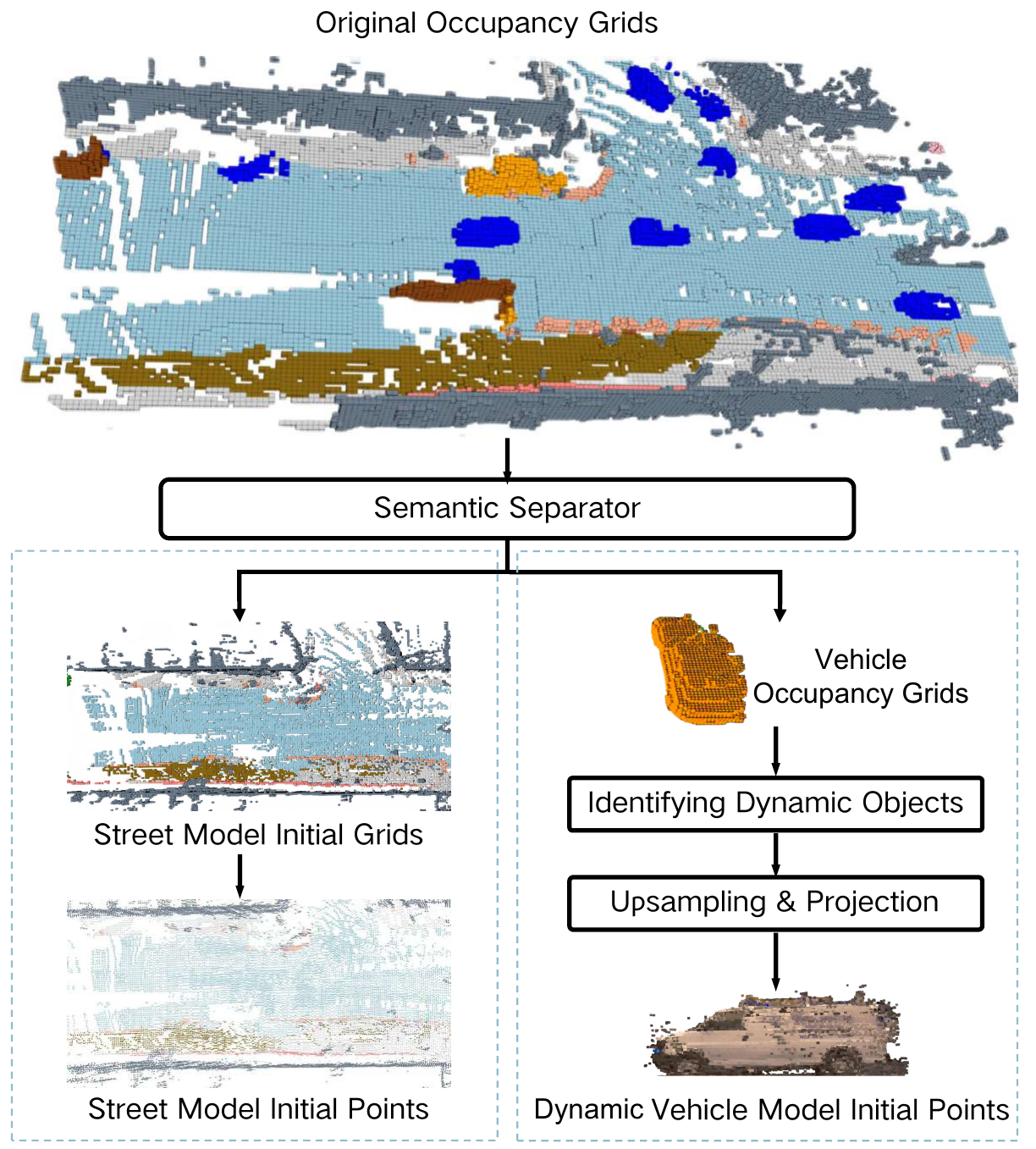} 
    \vspace{-10pt}
    \caption{\textbf{Initial point cloud generation process. }We extract dynamic vehicles from the street scene, then upsample and project them to obtain dense, colorized dynamic vehicle point cloud prior. The street scene can be converted into an initial background point clouds directly.} 
    \label{fig:dynamic} 
    \vspace{-19pt}
\end{figure}

In addition to the covariance and position matrices, each Gaussian contains a parameter $\alpha$ to represent opacity and a set of spherical harmonics coefficients (Eq.~\ref{6}) to represent the appearance of scenes. $l, m$ in Eq.~\ref{6} are referred to the degree and order that define the specific spherical harmonic function. In order to obtain color information of the original view, we also need to multiply the spherical harmonics coefficients with the spherical harmonics basis functions projected from the view direction. To get the semantic information for each Gaussian, we add logit $\beta_{s} \in \mathbb{R}^N$ to each point, where $N$ stands for the total number of semantic categories.

\begin{equation}
    \mathbf{z}_{s} = (z_{m, l})_{l:0\leq\ell\leq\ell_{max}}^{m:-\ell\leq m\leq\ell}
    \label{6}
\end{equation}

\noindent\textbf{Dynamic Vehicle model. }Autonomous driving scene contains multiple moving vehicles, and we also need to represent them with a set of optimizable point clouds. Observing dynamic vehicles from limited views, their movements result in significant changes in the surrounding space, so it's difficult to reconstruct them directly using 3DGS. We use the well-established detection and segmentation model~\cite{wu2019detectron2, cheng2021mask2former} to extract dynamic vehicle objects at the pixel level, and extract initial dynamic points cloud based on the semantic information of Occupancy grids, noting that the position of the point cloud is under the vehicle coordinate system.

The Gaussian properties of dynamic vehicles and streets are similar and they have the same meaning for the opacity $\alpha_{d}$ and the scale matrix $\mathbf{S}_d$. However, as we have mentioned above, their positions and rotations are under the vehicle coordinate system, which is different from the street scene. To avoid using the ground truth pose value, we represent the actual state of the dynamic vehicle by tracking its pose. The vehicle's pose can be represented by a set of rotation matrices $\mathbf{R}_t$ and translation vectors $\mathbf{T}_t$, as:

\begin{equation}
    \begin{aligned}
    &\boldsymbol{\mu}_w=\mathbf{R}_t\boldsymbol{\mu}_o+\mathbf{T}_t\\&\mathbf{R}_w=\mathbf{R}_o\mathbf{R}_t^T  
    \end{aligned}
    \label{7}
\end{equation}
Where $\boldsymbol{\mu}_w$ and $\mathbf{R}_w$ are the position matrix and rotation matrix of each Gaussian in the world coordinate system, $\boldsymbol{\mu}_o$ and $\mathbf{R}_o$ are the object position and rotation with respect to the car. From the prior knowledge we can get the covariance matrix of the dynamic vehicle by $\mathbf{R}_w$ and $\mathbf{S}_d$. In order to get more accurate vehicle pose, we replace the rotation matrix and position matrix of each frame as parameters like Eq.~\ref{8}, then we use them to get vehicle position and trajectory without ground truth of the vehicle trajectory.

\begin{equation}
\begin{aligned}
&\mathbf{R}_t^{\prime}=\mathbf{R}_t\Delta\mathbf{R}_t\\&\mathbf{T}_t^{\prime}=\mathbf{T}_t+\Delta\mathbf{T}_t
\end{aligned}
\label{8}
\end{equation}

The semantic representation of the dynamic vehicle model is also different from the street model, semantics in the street model is an N-dimensional vector (N is the number of semantic categories), the semantics of the vehicle model has only two categories for the vehicle semantic class (from the Occupancy prediction results) and non-vehicle class, so it's a one-dimensional scalar.

In the street model, we use spherical harmonics coefficients to represent the appearance of the scene. But when dealing with dynamic vehicles, their positions change with time. Therefore, it's wasteful to use multiple consecutive spherical harmonics coefficients to represent dynamic objects at each timestamp. Instead, we replace each SH coefficient $z_{m, l}$ with a set of Fourier transform coefficients $f\in\mathbb{R}^k$, constructing the 4D spherical harmonics coefficients (including the time dimension) in such a way that the real-valued inverse discrete Fourier transform can be performed to recover $z_{m, l}$ at a given time step.

\subsection{Occupancy Prior with Surrounding Views}

The original 3DGS generates sparse point clouds as prior with the help of structure-from-motion (SfM). In the task of reconstructing unbounded scenes of streets, it would be difficult to accurately represent dynamic objects and complex street scenes using SfM point clouds directly, this approach would produce some obvious geometric errors and incomplete recoveries. In order to provide accurate initialized point clouds for 3DGS, we convert the results predicted by ONet into an initialized point cloud to obtain accurate geometric information and maintain multi-camera consistency in the surround view alignment.

Specifically, we extract the vehicle point clouds based on the semantic information of OGs, and we define the vehicle location information for each timestamp as $\boldsymbol{\mu}_t$, if $\boldsymbol{\mu}_{t+1}-\boldsymbol{\mu}_t\geq\boldsymbol{\mu}_{th}$, we can label this vehicle as dynamic, where $\boldsymbol{\mu}_{th}$ denotes the threshold of the positional offset that determines it to be a dynamic object.

In order to generate densified point clouds to represent the dynamic vehicles, we upsample the point clouds of the dynamic objects with a voxel size of 0.05m. We project these points to the corresponding image planes and assign colors to them by querying pixel values. For each initial point $o$ of the dynamic vehicle, we transform its coordinates relative to the camera coordinate system, followed by the projection step described in Eq.~\ref{9}, where $x_p^q$ is the 2D pixel of the image, and $K \in \mathbb{R}^{3*3}$ is the internal reference matrix of each camera. $R_t$ and $T_t$ represent the orthogonal rotation matrix and translation vector.

\begin{equation}
    x_p^q=K[R_t\cdot o_d+T_t]
    \label{9}
\end{equation}


Finally, we convert the remaining Occupancy Grids into dense point clouds with position taken from their center coordinates. The specific process for generating initial point clouds of static and dynamic objects can be seen in Fig. ~\ref{fig:dynamic}. In addition to this, we also aggregate the dense point clouds with the point clouds generated by COLMAP in order to deal with distant buildings.

\subsection{Global Rendering via Gaussian Splatting}

To render the entire OG-Gaussian, we aggregate the contribution of each Gaussian to produce the final image. Previous methods represent scenes using neural fields, which require accounting for factors like lighting complexity when composing the scene. Our OG-Gaussian rendering approach is based on 3DGS, enabling high-fidelity rendering of autonomous driving scenes by projecting the Gaussians of all point clouds into 2D image space.

Given a rendering timestamp \(t\), we first calculate the spherical harmonics coefficients using Eq.~\ref{6}. After transforming point clouds from the vehicle coordinate system to the world coordinate system, we combine the street model and the dynamic model into a global model. Using the camera's extrinsic parameters \(W\) and intrinsic parameters \(K\), we project point clouds onto a 2D plane and calculate each point's parameters in the 2D space. In the Eq.~\ref{10}, \(J\) is the Jacobian matrix of \(K\), while \(\mu'\) and \(\Sigma'\) represent the position and covariance matrix in the 2D image space.

\begin{equation}    
\begin{aligned}
\mu^{\prime}&=\mathbf{KW}\boldsymbol{\mu}\\\Sigma^{\prime}&=\mathbf{JW}\Sigma\mathbf{W}^T\mathbf{J}^T
\end{aligned}
\label{10}
\end{equation}

After this, we can calculate the appearance color \(a\) of each pixel based on the opacity of the points. In Eq.~\ref{11}, \(\alpha_i\) is the product of opacity \(\alpha\) and the probability of the 2D Gaussian, while \(a_i\) is the color derived from the spherical harmonic \(z\) with a specific view direction.

\begin{equation}
\begin{aligned}
\mathbf{a}=\sum_{i\in N}\mathbf{a}_i\alpha_i\prod_{j=1}^{i-1}(1-\alpha_j) 
\end{aligned}
\label{11}
\end{equation}

\section{EXPERIMENT}

\begin{table}[b]
    \centering \vspace{-2em}
    \caption{\textnormal{\textbf{Overview of the scene. }}}
    \vspace{-10pt}
    \begin{tabular}{ccccc}
    \toprule
         Scene & Weather & Lighting & Area & Traffic flow \\
    \midrule
         00 & Rainy & dark & urban & low\\
         01 & Cloudy & ordinary & suburbia & high \\
         02 & Sunny & bright & downtown & middle\\        
    \bottomrule
    \end{tabular}
    \label{tab:scene}
\end{table}

\begin{table*}[t]
\vspace{+6pt}
\centering
\caption{\textnormal{\textbf{Comparison of overall performance between OG-Gaussian and baselines on the Waymo dataset} Here, PSNR-dy refers to the PSNR value specifically for rendering dynamic vehicles. The result with bold black font is the best result for this scene.}}
\vspace{-10pt}
\begin{tabular}{lcccccccc}
\toprule
Methods & Inputs & Modal & Scene & PSNR[dB]~$\uparrow$ & PSNR-dym[dB]~$\uparrow$ & SSIM~$\uparrow$ & LPIPS~$\downarrow$ & FPS~$\uparrow$ \\  
\midrule
\multirow{3}{*}{MARS~\cite{wu2023mars}} & \multirow{3}{*}{Images} & \multirow{3}{*}{Camera} & 00 & 31.07 & 21.24 & 0.901 & 0.243 & 0.67\\
 &  & & 01 & 30.11 & 20.96 & 0.906 & 0.251 & 0.62\\
 &  & & 02 & 31.54 & 21.63 & 0.917 & 0.239 & 0.73\\
 \hline
\multirow{3}{*}{EmerNeRF~\cite{yang2023emernerf}}  & \multirow{3}{*}{Images+LiDAR points} & \multirow{3}{*}{Camera+LiDAR} & 00 & 33.24 & 23.21 & 0.911 & 0.221 & 1.94\\
 &  & & 01 & 31.18 & 22.86 & 0.904 & 0.226 & 1.62\\
 &  & & 02 & 33.07 & 23.09 & 0.916 & 0.241 & 1.96\\
\hline
\multirow{3}{*}{3DGS~\cite{kerbl20233d}} & \multirow{3}{*}{Images+SfM points} & \multirow{3}{*}{Camera}& 00 & 30.21 & / & 0.916 & 0.181 & \textbf{209}\\
 &  & & 01 & 29.16 & / & 0.907 & 0.192 & \textbf{204}\\
 &  & & 02 & 32.46 & / & 0.923 & 0.179 & \textbf{218}\\
\hline
\multirow{3}{*}{Street Gaussian~\cite{yan2024street}} & \multirow{3}{*}{Images+LiDAR points} & \multirow{3}{*}{Camera+LiDAR} & 00 & \textbf{35.92} & 24.96 & 0.961 & 0.096 & 133\\
 &  & & 01 & 32.91 & \textbf{24.53} & 0.943 & \textbf{0.094} & 129\\
 &  & & 02 & 36.14 & 26.14 & 0.959 & 0.091 & 137\\
\hline
\multirow{3}{*}{Ours} & \multirow{3}{*}{Images} & \multirow{3}{*}{Camera} & 00 & 35.87 & \textbf{25.17} & \textbf{0.965} & \textbf{0.093} & 143\\
 &  & & 01 & \textbf{33.24} & 24.28 & \textbf{0.958} & 0.097 & 139\\
 &  & & 02 & \textbf{36.28} & \textbf{26.33} & \textbf{0.962} & \textbf{0.089} & 148\\
\bottomrule
\end{tabular}
\vspace{-5pt}
\label{tab:overall}
\end{table*}

\subsection{Experiment Setups}
We perform 30, 000 iterations on OG-Gaussian, using the same learning rate settings as the original 3DGS. The learning rates for the rotational transformation $\Delta R_t$ and translational transformation $\Delta T_t$ are set to 0.001 and 0.005. The Adam optimizer is used for optimization. All experiments are conducted on a machine equipped with an Nvidia A100 GPU.


\noindent \textbf{Dataset. } The Waymo Open Dataset~\cite{sun_2020_cvpr} is widely used in autonomous driving research due to its high-quality 360-degree panoramic imagery, which offers comprehensive coverage of the vehicle's surrounding\cite{10.1145/3636534.3649386}. The dataset's quality, diversity, and scale make it a valuable asset for researchers and developers in autonomous driving. It has been widely adopted due to its realistic and challenging dataset, which enables the development and testing of algorithms in conditions that closely resemble real-world driving. 

\begin{figure*}[!h]
\centering
\includegraphics[width=0.94\textwidth]{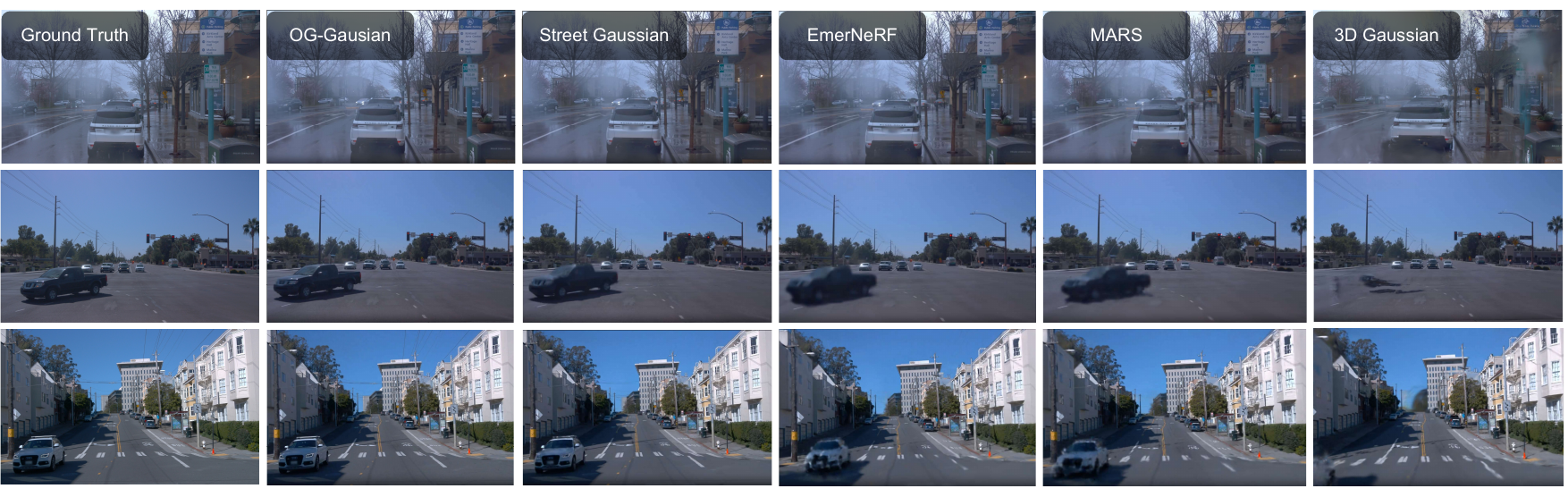} \vspace{-1em}
\caption{\textbf{Qualitative comparison of different reconstruction methods.} Each column in the figure represents the qualitative results of different reconstruction methods, with each row showing the same viewpoint.}
\label{fig:experiment} 
\vspace{-15pt}
\end{figure*}

\noindent\textbf{Baselines. }For selecting baselines, we choose several representative outdoor scene reconstruction methods, including both implicit and explicit representations. We will introduce these methods in detail below.
\begin{itemize}
    \item MARS~\cite{wu2023mars} is an autonomous driving simulator using NeRF to independently model dynamic objects and static backgrounds.
    \item EmerNeRF~\cite{yang2023emernerf} enhances dynamic scene reconstruction through self-supervised separation of scene elements and a flow field that aggregates multi-frame features.
    \item 3DGS~\cite{kerbl20233d} uses Gaussian ellipsoids for high-quality 3D scene representation, excelling in rendering speed.
    \item Street Gaussians~\cite{yan2024street} extends 3DGS for urban scenes, combining LiDAR point clouds and dynamic object segmentation for view synthesis at impressive speeds.
\end{itemize}

\subsection{Comparisons with Baselines}

In terms of experimental scene settings, we choose three representative sequences from the Waymo Open Dataset~\cite{sun_2020_cvpr}. These sequences vary in weather conditions, traffic flow and street types as shown in Tab.~\ref{tab:scene}, allowing us to evaluate the reconstruction performance across different scenarios.

Tab.~\ref{tab:overall} presents a quantitative comparison of our method with the baselines in terms of rendering quality and speed. We use Peak Signal-to-Noise Ratio(PSNR), Structural Similarity Index Measure(SSIM)~\cite{wang2004image} and Learned Perceptual Image Patch Similarity(LPIPS)~\cite{zhang2018unreasonable} as metrics to assess the reconstruction quality. To specifically evaluate the reconstruction of dynamic vehicles, we render the objects identified as dynamic and use their PSNR (PSNR-dym) as the metric to measure the effect of dynamic object reconstruction. The experiment results show that the average PSNR for Mars~\cite{wu2023mars} and EmerNeRF~\cite{yang2023emernerf}, dynamic-static separation reconstruction methods based on NeRF, are 30.91 and 32.50, respectively. In comparison, our method achieves a higher average of 35.13. Regarding rendering speed, our method is two orders of magnitude faster than the aforementioned approaches. Street Gaussian~\cite{yan2024street}, which is based on LiDAR point clouds, performs similarly to ours in both rendering quality and speed. But this method requires expensive LiDAR data, while our method operates in a purely visual modality. The original 3DGS~\cite{kerbl20233d} method is faster than ours, but it does not separately reconstruct dynamic objects, which result in blurry artifacts in regions involving dynamic objects. As a consequence, it only achieved a PSNR of 30.61 and an SSIM of 0.954, since it does not separate dynamic objects, we don't calculate its PSNR-dym. We visualized each reconstruction method across three sets of scenes to evaluate their qualitative performance, as shown in Fig.~\ref{fig:experiment}.

\begin{table}[b]
\centering \vspace{-15pt}
\caption{\textnormal{\textbf{Ablation study results. }The results of each experimental setup.}}
\vspace{-1em}
\begin{tabular}{lcccc}
\toprule
Initial Method & PSNR~$\uparrow$ & PSNR-dym~$\uparrow$ & SSIM~$\uparrow$ & LPIPS~$\downarrow$ \\ 
\midrule
w/o OG,SfM & 26.01   & 19.08   & 0.895   & 0.247   \\
w/o OG   & 32.62   & 19.94  & 0.923  & 0.180  \\
Ours  & \textbf{35.13} & \textbf{25.26} & \textbf{0.962} & \textbf{0.093}  \\
\bottomrule
\end{tabular}
\label{tab:ablation}
\end{table}

\subsection{Ablations and Analysis}
To validate the significance of OG-based point clouds as prior, we designed the following ablation experiment. Three experimental setups were configured, all sharing identical settings except for the point clouds prior. In the first setup, we use the point clouds combined by OGs and SfM points from COLMAP as prior. In the second setup, only the SfM point clouds are used as prior, while in the third setup, randomly generated point clouds are employed as prior.

\begin{figure}[h]
    \centering \vspace{-10pt}
    \includegraphics[width=0.9\columnwidth]{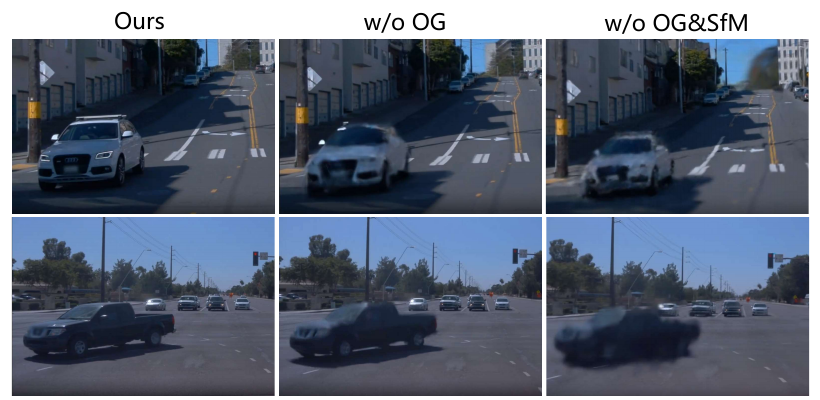} 
    \vspace{-0.8em}
    \caption{\textbf{Visual ablation results for specific scenes.} The results of different initialization methods.} 
    \label{fig:ablation} 
    \vspace{-2pt}
\end{figure}

Tab.~\ref{tab:ablation} shows that adding the Occupancy Grid improves the overall PSNR of the reconstructed scene (35.13) compared to using only the SFM point cloud (32.62), and PSNR-dym increases significantly. On the other hand, using randomly generated point clouds based on spatial size performs worse than the other two methods. This experiment indicate the importance of initial point clouds in 3D scene reconstruction. The qualitative comparison of different initialization methods can be seen in Fig. ~\ref{fig:ablation}. Here, each metric represents the average value across the three experimental scenarios.

\section{CONCLUSIONS}
We introduce OG-Gaussian, an efficient method that incorporates the Occupancy Grids into 3DGS for reconstructing outdoor autonomous driving scenes. Our approach leverages the prior provided by the Occupancy Grid for scene reconstruction while also separating and reconstructing dynamic vehicles from static street scenes. We achieve performance on par with LiDAR-dependent SOTA while relying solely on camera images. Our method will enable future researchers to reconstruct their autonomous driving scenes quickly and at a low cost, contributing to the advancement of autonomous driving technology.

\section*{acknowledgment} This work is supported by the National Natural Science Foundation of China (No. 62332016) and the Chinese Academy of Sciences Frontier Science Key Research Project ZDBS-LY-JSC001.

\bibliographystyle{IEEEtran}
\bibliography{IEEEabrv, reference}

\end{document}